\documentclass[11pt,letterpaper]{article}
\usepackage{naaclhlt2016}
\usepackage{times}
\usepackage{helvet}
\usepackage{courier}
\usepackage{latexsym}
\usepackage{url}
\usepackage{graphicx}
\usepackage{multirow}
\usepackage{amsmath}

\naaclfinalcopy 

\newcounter{notecounter}
\newcommand{\enotesoff}{\long\gdef\enote##1##2{}}
\newcommand{\enoteson}{\long\gdef\enote##1##2{{
\stepcounter{notecounter}
\large\bf
\hspace{1cm}\arabic{notecounter} $<<<$ ##1: ##2
$>>>$\hspace{1cm}}}}
\enoteson
\enotesoff

\def\tabref#1{Table~\ref{tab:#1}}
\def\tablabel#1{\label{tab:#1}\label{p:#1}}

\def\eqref#1{Eq.~\ref{eqn:#1}}

\begin{document}

  \title{Impact of Coreference Resolution on Slot Filling}

 \author{Heike Adel \and Hinrich Sch\"{u}tze \\
         Center for Information and Language Processing (CIS)\\
         LMU Munich, Germany\\
         {\tt heike.adel@cis.lmu.de}}

\date{}
 
\maketitle

\begin{abstract}
In this paper, we demonstrate the importance of coreference resolution 
for natural language processing on the example of the TAC Slot Filling
shared task. We illustrate the strengths and weaknesses of automatic coreference
resolution systems and provide experimental results to show
that they improve performance in the slot filling end-to-end setting.
Finally, we publish \texttt{KBPchains}, a resource containing 
automatically extracted coreference chains from
the TAC source corpus in order to support other researchers working on this topic.
\end{abstract}

\section{Introduction}
Coreference resolution systems group noun phrases (mentions) that
refer to the same entity into the same chain.
Mentions can be full names (e.g., John Miller), pronouns (e.g., he), 
demonstratives (e.g., this), comparatives (e.g., the first)
or descriptions of the entity (e.g. the 40-year-old) \cite{coreference08}.
Although coreference resolution has been a research focus for
several years, systems are still far away from being perfect.
Nevertheless, there are many tasks in natural language
processing (NLP) which would benefit from coreference information, such as
information extraction, question answering or summarization \cite{coreference13}.
In \cite{adelEmnlp2014}, for example, we showed that coreference
information can also be incorporated into word embedding training.
In general, coreference resolution systems can be used
as a pre-processing step or as a part of a pipeline of different modules.

Slot Filling is an information extraction task 
which has become popular in the last years
\cite{sfTask2014}. It is a shared task organized by 
the Text Analysis Conference (TAC).
The task aims at extracting information about persons, organizations
or geo-political entities from a large collection of news, web 
and discussion forum documents.
An example is ``Steve Jobs'' for the slot ``X founded Apple''.
Thinking of a text passage like ``Steve Jobs was an American businessman.
In 1976, he co-founded Apple'', it is clear that coreference resolution
can play an important role for finding the correct slot filler value.

In this study, we investigate
how coreference resolution could help to improve performance
on slot filling and which challenges exist.
Furthermore, we present how we pre-processed the TAC
source corpus with a coreference resolution system in order to be able
to run the slot filling system more efficiently. In addition to this paper,
we also publish the results of this pre-processing since it
required long computation time and much 
resources.\footnote{\texttt{cistern.cis.lmu.de}}

\begin{figure*}
\centering
\includegraphics[width=.7\textwidth]{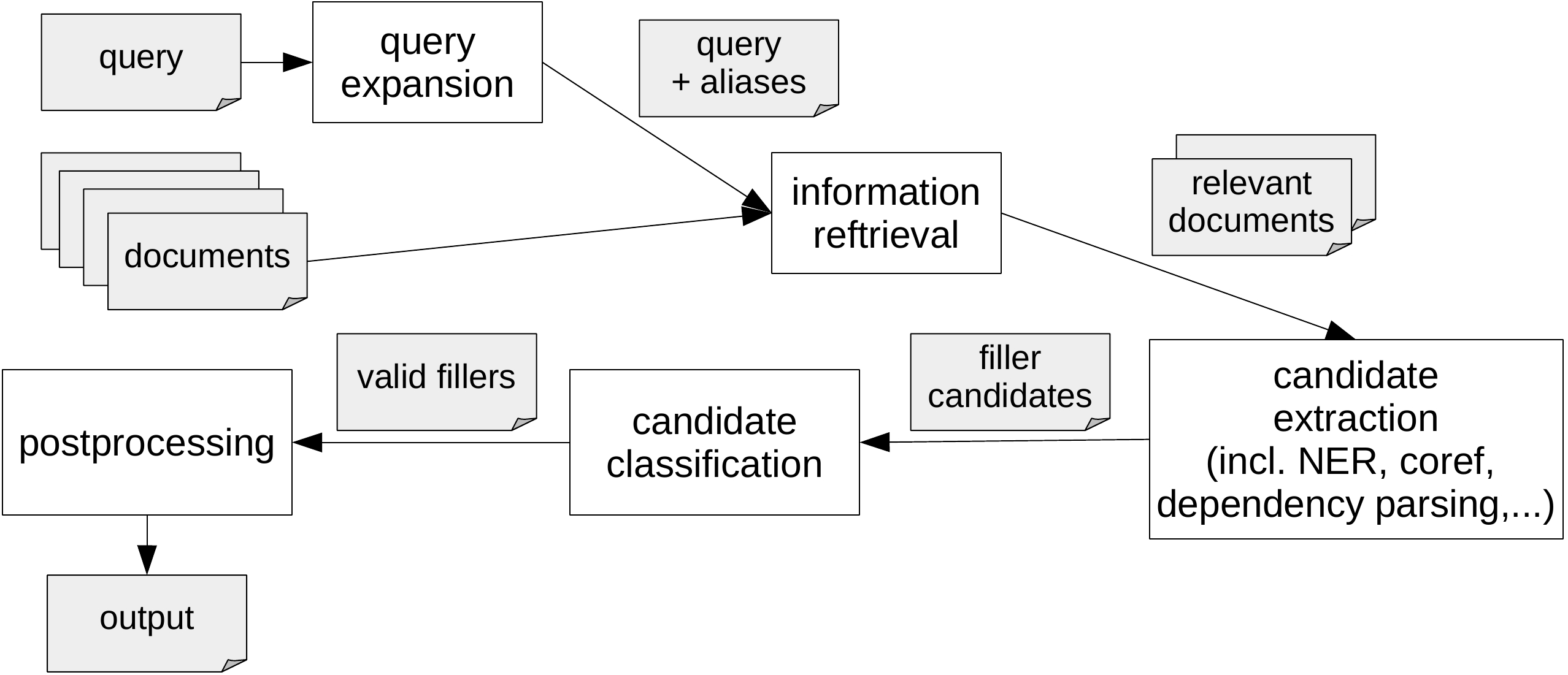}
\caption{Typical slot filling pipeline}
\label{fig:sys}
\end{figure*}

\section{Related work}
The slot filling task has been organized since 2009.
The top ranked systems of the last years achieved F1 scores of 37.28 (2013) \cite{roth2013},
36.77 (2014) \cite{stanford2014} and 31.48 (2015).
In 2015, the task has been merged with the Cold Start track of the
same conference. This led to several changes in the number of relations,
the evaluation documents and the outputs expected from the systems \cite{sfTask2015}.

Previous studies and error analyses have shown that coreference resolution
is an important component to increase the recall of
slot filling systems \cite{sfApproaches,analysis2012,analysisRecall}.
\newcite{analysis2012} identified coreference failures
as the second most frequent error source of slot filling systems 
(after inference failures). In most cases, nominal anaphors were
not resolved correctly.
\newcite{analysisRecall} investigated possible causes of recall loss
in a slot filling system. They described that coreference resolution
provided higher recall but might be inefficient since it requires
a lot of time and resources. Moreover, they argued
that the overall results of a slot filling system might be
better without coreference resolution since it can have
a negative impact on precision.
In contrast, our experiments in this study show that the increased number
of true positives when using coreference resolution 
has a much higher impact on the final results. For coping with
the problem of time-consuming coreference resolution, we
prepared and publish \texttt{KBPchains}, a coreference resource for slot filling.

\section{Slot filling task}
The main idea of slot filling is to extend a knowledge base by
extracting pre-defined relations between (named) entities 
from text data. Systems are provided with a large collection of 
text documents and a query file including entities 
and the relations to find in the text. As output,
they have to provide the second argument for each relation.
For entity ``Apple'' and relation ``org:founded\_by'', for example,
the systems need to extract ``Steve Jobs'', ``Steve Wozniak'' 
and ``Ronald Wayne'' along with text passages for justification.

This task combines several NLP challenges
like information retrieval, information extraction,
relation classification and knowledge inference.
Until 2014, the slot filling shared task
included 41 relations (25 for persons and 16 for organizations) \cite{sfTask2014}.
Since 2015, these relations have been extended to all possible
inverse relations which introduced a new query entity
type (geo-political entity) and augmented 
the set of relations to 64 (27 for persons,
20 for organizations and 17 for geo-political entities) \cite{sfTask2015}.
Table \ref{tab:exRel} provides exemplary relations for the different
entity types.

The input for a slot filling system is an xml query
containing the name and type of the entity, an exemplary
occurence of the entity in the document collection
and the slot to be filled.
The expected output of the system contains, i.a.,
a provenance for the slot filler in the document 
collection, the slot filler itself,
the type of the filler ($\in{PER, ORG, GPE, STRING}$),
its offsets in the document collection, 
as well as a confidence value of the system.

The document collection from which the slot fillers should be extracted
is quite large: until 2014, it consisted of about 2.1 million documents,
in 2015 the number was reduced to about 50,000 documents. The documents
comprise newswire, web and discussion forum texts.
Therefore, the slot filling task is more than relation extraction
for pre-defined relations: It also includes challenges like information retrieval and 
coping with different genres.

Most slot filling systems are a pipeline of different components,
such as query expansion, information retrieval, candidate extraction,
candidate classification and postprocessing. Figure \ref{fig:sys}
depicts a typical system.
We performed a detailed analysis of the errors of these
components and found that one of the most important sources of error
is failure of coreference resolution in the candidate extraction step.

\section{Coreference resolution for slot filling}
In our study, we have identified two main reasons 
why coreference resolution can improve slot filling performance.
The first reason is that both arguments of a relation can be
pronouns referring to the entity or filler in question.
Consider the relation ``per:parents'' and the sentence
``Bill is the father of Jane.'' Both entities ``Bill'' and
``Jane'' might have been mentioned in sentences before
and could now be replaced by pronouns:
``He is the father of Jane'',
``Bill is her father'' or ``He is her father''.
If a slot filling system only extracts sentences
with appearances of the full name of a person, it might
miss many relevant sentences which can reduce the recall of
the whole system drastically.
As \newcite{analysisRecall} pointed out, the recall
losses cannot be recovered by subsequent pipeline modules.

\begin{table}
\footnotesize
\centering
\begin{tabular}{l|l}
 Entity type & exemplary slots\\
 \hline
 Person & age, title, date of birth,\\
 & schools attended\\
 \hline
 Organization & employees, founders, \\
 & city of headquarters\\
 \hline
 Geo-political entity & births in country, \\
 & residents in city
\end{tabular}
\caption{Exemplary relations for slot filling}
\label{tab:exRel}
\end{table}

The second reason is that coreference resolution can provide
slot fillers ``for free'': If a phrase like ``The Hawaii-born''
is coreferent to the entity in question,
it not only provides an additional sentence with information
about the entity but also directly the location of birth
(without the need of classification).
Similar phrases can provide the age, a title
or the religion of a person or the location of
headquarters of an organization.

\subsection{Coreference resource}
As motivated above, coreference information is a very important resource
for participants of the slot filling task
or related knowledge base population tasks on the same documents. Since we found that
the coreference resolution component is one of the bottlenecks
which considerably slows down our slot filling pipeline, we
have pre-processed the TAC source corpus by tagging its documents 
using Stanford CoreNLP \cite{coreNLP}. We call this resource
of coreference chains \texttt{KBPchains} and share it
(in the form of document-offset spans) 
on our website.\footnote{\texttt{cistern.cis.lmu.de}}
Although CoreNLP is publicly available, \texttt{KBPchains} will save researchers
much time and resources (cf., \newcite{analysisRecall} who 
mentioned the need for efficient coreference resolution when
processing the large slot filling corpora).
Table \ref{stat} lists statistics about the extracted coreference chains and
their mentions. In addition to the minimum, maximum, average and median numbers
of chains per document, mentions per chain and words per mention, we 
also report the number of mentions which are pronouns, the number
of singletons (chains consisting of only one mention) and the number
of chains with only identical mentions.

\begin{table}[h!]
\centering
\footnotesize
 \begin{tabular}{l|r}
  Total number of chains & 53,929,525\\
  Chains per document: min & 0 \\
  Chains per document: max & 2061\\
  Chains per document: avg & 26.18\\
  Chains per document: median & 15\\
  \hline
  Total number of mentions & 197,566,321\\
  Mentions per chain: min & 1 \\
  Mentions per chain: max & 3428 \\
  Mentions per chain: avg & 3.66\\
  Mentions per chain: median & 2 \\
  \hline
  Words per mention: min & 1\\
  Words per mention: max & 900\\
  Words per mention: avg & 3.05\\
  Words per mention: median & 2\\
  \hline
  \# pronoun mentions & 51,139,283\\
  \hline
  \# singletons & 13,189\\
  \hline
  \# chains with identical mentions & 16,016,935
 \end{tabular}
 \caption{Statistics of \texttt{KBPchains}}
 \label{stat}
\end{table}

\section{Analysis of coreference resolution errors}
\label{sec:analysis}
Coreference resolution systems produce acceptable results 
but are still far away from being perfect.
In an analysis of the results of Stanford CoreNLP
on the TAC source corpus in the context of our slot filling system,
we found the following flaws being most prominent:
Wrongly linked pronoun chains, unlinked pronoun chains
and no recognition of coreferent phrases like ``the 42-year-old'',
``the author'' or ``the California-based company''.
In the following, we describe the effect of these failures on
the slot filling system.

\textbf{Wronly linked pronoun chains.} 
If a pronoun chain is 
wrongly linked to the entity in question, all sentences with
pronouns of this chain will be extracted as sentences containing
information about the entity. This increases the number of
falsely extracted sentences and as a result also the number
of possible filler candidates. All those false positive
filler candidates will be passed to the candidate evaluation 
module and can easily lead to a lower precision in the final output.
(Either because the candidate evaluation makes a wrong decision, too
or because -- in the worst case -- the relation in question holds
between the pronoun and the filler candidate but not between the
entity in question and the filler candidate.)

\textbf{Unlinked pronoun chains.}
If a coreference chain consists of only pronouns without any entity mention,
the slot filling system cannot decide to which entity it belongs to
and will omit it. If the pronouns of the chain are 
coreferent to the entity in question, the chance that
the slot filling system misses information which are relevant
to the slot in question is quite high. As a result, the
recall of the end-to-end system will be reduced.
A solution to this problem could be a post-processing
of these unlinked pronoun chains, a challenge we will
investigate in the future.

\textbf{No recognition of nominal anaphors.}
Phrases like ``the 42-year-old'' or ``the California-based company'' 
may occur directly after a sentence
with the entity in question but are often not recognized as being
coreferent to it. However, if they refer to this entity,
they first contain possibly relevant information 
(like the age of a person).
Second, the sentence in which they appear could mention
additional information about the entity.
Omitting these sentences and these phrases can therefore
reduce the recall of the slot filling system.
In our system, we cope with these cases by 
explicitely looking for such phrases in the sentence
following a mention of the entity in question.

\textbf{Additional findings.}
We perform a manual analysis of the extracted
coreference chains in ten randomly chosen documents
with the following results.
\begin{enumerate}
\item We found that HTML tags in documents lead to errors in coreference chains
(e.g. persons and locations are clustered with HTML tags in the same
chain). 
\item Especially in web documents, the extracted mentions are often too long (containing
not only the coreferent entity but additional text, such as ``Markus , the error'').
In some of these cases, errors from previous sentence splitting steps propagate:
There are also overlong mentions spanning the boundary of a sentence.
\item Furthermore, we found many cases of unlinked chains (e.g. [``they'', ``they''])
or wrongly linked chains (e.g. [``Hamas and Fatah movement'', ``his movement'']
or [``her'', ``I''] with ``her'' and ``I'' referring to different entities).
\item While some chains correctly cover all the pronouns referring to an entity, other
chains do not include all relevant pronouns of the text.
\item Correct and clean chains are often found in the context of dates ([``Nov 20, 2009'', 
``November 20 next year'', ``next year'']) or recurring identical mentions of events.
\end{enumerate}

\section{Experiments with end-to-end system}
In order to investigate the impact of coreference resolution
on slot filling empirically, we perform end-to-end experiments on
the TAC evaluation data from 2015. Our system with coreference resolution
was one of the top-performing systems in the official evaluations 2015 \cite{cis2015}.
It follows the pipeline shown in Figure \ref{fig:sys}. For a more detailed
descriptions of its component, see \cite{cis2015}.
\tabref{ch:slotfilling-tab:ablationCoref} shows its results
with (+) and without (-) coreference resolution in the candidate extraction component.

\begin{table}
\centering
\begin{tabular}{lrrrr}
   coreference & P & R & F1 & $\Delta$F1\\
   \hline
 + & 31.67 & 23.97 & 27.29 &\\
 - & 19.33 & 22.40 & 20.75 & -6.54\\
   \end{tabular}
   \caption{Impact of coreference resolution on performance on hop 0 of 2015 evaluation data}
\tablabel{ch:slotfilling-tab:ablationCoref}
\end{table}

The number of true positives is reduced considerably (from 361 to 321)
when the system does not use coreference information.
The number of false positives is also lower, but the final results show
that the impact of the number of true positives is larger since
it affects both precision and recall: The F1 score
drops by more than 6 points when omitting coreference resolution.

To conclude, in order to provide the classification and
postprocessing modules with a recall as high as possible, coreference resolution
is a crucial part of the system. Despite of the errors identified in
Section \ref{sec:analysis}, an automatic coreference system 
still performs well enough to improve
the performance on slot filling.

\section{Conclusion}
In this work, we analyzed the impact of coreference resolution on the NLP task
slot filling. We showed that coreference information improves the slot
filling system performance and outlined the most important challenges we have discovered
in an analysis of coreference resolution errors.
Since the TAC source corpus is very large, we will publish 
\texttt{KBPchains}, a resource containing the coreference chains
which we have extracted automatically.

\section*{Acknowledgments}
Heike  Adel  is  a  recipient  of  the  Google  European
Doctoral Fellowship in Natural Language Processing 
and this research is supported by this fellowship.
This work was also supported by DFG (grant SCHU
2246/4-2).

\bibliographystyle{naaclhlt2016}

\bibliography{refs}

\begin{thebibliography}{}

\bibitem[\protect\citename{Adel and Sch{\"u}tze}2014]{adelEmnlp2014}
Heike Adel and Hinrich Sch{\"u}tze.
\newblock 2014.
\newblock Using mined coreference chains as a resource for a semantic task.
\newblock In {\em Proceedings of the 2014 Conference on Empirical Methods in
  Natural Language Processing (EMNLP)}, pages 1447--1452. Association for
  Computational Linguistics.

\bibitem[\protect\citename{Adel and Sch\"{u}tze}2015]{cis2015}
Heike Adel and Hinrich Sch\"{u}tze.
\newblock 2015.
\newblock Cis at tac cold start 2015: Neural networks and coreference
  resolution for slot filling.
\newblock In {\em Proceedings of Text Analysis Conference (TAC)}.

\bibitem[\protect\citename{Angeli \bgroup et al.\egroup }2014]{stanford2014}
Gabor Angeli, Sonal Gupta, Melvin~Johnson Premkumar, Chris Manning, Chris Re,
  Julie Tibshirani, Jean~Y. Wu, Sen Wu, and Ce~Zhang.
\newblock 2014.
\newblock Stanford's distantly supervised slot filling systems for {KBP} 2014.
\newblock In {\em Proceedings of the Text Analysis Conference (TAC)}.

\bibitem[\protect\citename{Clark and Gonz{\'a}lez-Brenes}2008]{coreference08}
Jonathan~H Clark and Jos{\'e}~P Gonz{\'a}lez-Brenes.
\newblock 2008.
\newblock Coreference resolution: Current trends and future directions.
\newblock {\em Language and Statistics II Literature Review}, pages 1--14.

\bibitem[\protect\citename{Ji and Grishman}2011]{sfApproaches}
Heng Ji and Ralph Grishman.
\newblock 2011.
\newblock Knowledge base population: Successful approaches and challenges.
\newblock In {\em Proceedings of the 49th Annual Meeting of the Association for
  Computational Linguistics: Human Language Technologies}, pages 1148--1158,
  Portland, Oregon, USA. Association for Computational Linguistics.

\bibitem[\protect\citename{Lee \bgroup et al.\egroup }2013]{coreference13}
Heeyoung Lee, Angel Chang, Yves Peirsman, Nathanael Chambers, Mihai Surdeanu,
  and Dan Jurafsky.
\newblock 2013.
\newblock Deterministic coreference resolution based on entity-centric,
  precision-ranked rules.
\newblock {\em Computational Linguistics}, 39(4):885--916.

\bibitem[\protect\citename{Manning \bgroup et al.\egroup }2014]{coreNLP}
Christopher~D. Manning, Mihai Surdeanu, John Bauer, Jenny Finkel, Steven~J.
  Bethard, and David McClosky.
\newblock 2014.
\newblock The {Stanford} {CoreNLP} natural language processing toolkit.
\newblock In {\em Proceedings of 52nd Annual Meeting of the Association for
  Computational Linguistics: System Demonstrations}, pages 55--60.

\bibitem[\protect\citename{Min and Grishman}2012]{analysis2012}
Bonan Min and Ralph Grishman.
\newblock 2012.
\newblock Challenges in the knowledge base population slot filling task.
\newblock In {\em Proceedings of the Language Resources and Evaluation
  Conference (LREC)}, pages 1137--1142.

\bibitem[\protect\citename{Pink \bgroup et al.\egroup }2014]{analysisRecall}
Glen Pink, Joel Nothman, and James~R Curran.
\newblock 2014.
\newblock Analysing recall loss in named entity slot filling.
\newblock In {\em Proceedings of Empirical Methods in Natural Language
  Processing (EMNLP)}, pages 820--830.

\bibitem[\protect\citename{Roth \bgroup et al.\egroup }2013]{roth2013}
Benjamin Roth, Tassilo Barth, Michael Wiegand, Mittul Singh, and Dietrich
  Klakow.
\newblock 2013.
\newblock Effective slot filling based on shallow distant supervision methods.
\newblock In {\em Proceedings of the Text Analysis Conference (TAC)}.

\bibitem[\protect\citename{TAC}2014]{sfTask2014}
TAC.
\newblock 2014.
\newblock Task description for {E}nglish slot filling at {TAC KBP} 2014.
\newblock
  \url{http://surdeanu.info/kbp2014/KBP2014_TaskDefinition_EnglishSlotFilling_1.1.pdf}.

\bibitem[\protect\citename{TAC}2015]{sfTask2015}
TAC.
\newblock 2015.
\newblock Cold start knowledge base population at {TAC} 2015: Task description.
\newblock
  \url{http://www.nist.gov/tac/2015/KBP/ColdStart/guidelines/TAC_KBP_2015_ColdStartTaskDescription_1.1.pdf}.

\end{thebibliography}

\end{document}